\documentclass{article}

\usepackage{microtype}
\usepackage{graphicx}
\usepackage{subcaption}
\usepackage{booktabs} % for professional tables

\usepackage{hyperref}

\usepackage[accepted]{icml2026}

\usepackage{amsmath}
\usepackage{amssymb}
\usepackage{mathtools}
\usepackage{amsthm}

\usepackage[capitalize,noabbrev]{cleveref}

\theoremstyle{plain}

\theoremstyle{definition}

\theoremstyle{remark}

\usepackage[textsize=tiny]{todonotes}

\icmltitlerunning{``AI is (not) the new...": A Diagnostic Analogy Framework for Generative AI's Cultural Impacts}

\begin{document}

\twocolumn[
  \icmltitle{``AI is (not) the new..": A Diagnostic Analogy Framework for Generative AI's Cultural Impacts}



  \icmlsetsymbol{equal}{*}

  \begin{icmlauthorlist}
    \icmlauthor{Rida Qadri}{Google}
    \icmlauthor{Vinodkumar Prabhakaran}{Google}
    \icmlauthor{Remi Denton}{Google}
  \end{icmlauthorlist}

  \icmlaffiliation{Google}{Google Research}

  \icmlcorrespondingauthor{Rida Qadri}{ridaqadri@google.com}

  \icmlkeywords{Machine Learning, ICML}

  \vskip 0.3in
]



\printAffiliationsAndNotice{}  % no special notice (required even if empty)

\begin{abstract}
Generative AI is reshaping the cultural infrastructures through which knowledge is found, synthesized, and held accountable. To make sense of this shift, scholars and policymakers reach for historical analogies of technologies such as the printing press, steam power or electricity. But these comparisons are typically imprecise about which property of the technology carries the comparison, and imprecise analogies produce imprecise governance by designing interventions against the wrong property of the system. This paper offers a diagnostic framework for analyzing how generative AI can transform epistemic and cultural practice. This paper offers a diagnostic framework for analyzing how generative AI can transform epistemic and cultural practice. We decompose each intervention into three coordinates:  the epistemic site at which a technology acts, the governing logic by which it organizes its object, and the technical mechanism through which the logic is instantiated. This framework allows us to distinguish between structural cultural consequences, which follow from the mechanism itself, from contingent ones, which remain open to design and institutional choice. Applying the framework to information discovery and knowledge synthesis, we show how the shift from indexicality to inference and from editorial authority to statistical consensus produces specific, traceable cultural effects  and reveals governance levers that gestalt analogy obscures.
\end{abstract}

\section{Introduction}

Generative artificial intelligence (GenAI) is rapidly restructuring the epistemic infrastructures through which human knowledge is discovered, synthesized, and held accountable \cite{Farrelletal2025}. As scholars, technologists, and policymakers struggle to comprehend the scale of this transition, they have routinely reached for historical anchors. \cite{Farrelletal2025}, framing GenAI variously as a new printing press \cite{Smith2025,Wheeler2024}, electricity \cite{ng2017electricity} or steam power \cite{Simosetal2022}. Such analogical comparisons are not merely rhetorical but serve as conceptual infrastructure: by creating frames of understanding and legitimacy they can structure how we understand a technology,  which historical lessons are deemed relevant,   and which societal harms register as foreseeable \cite{Petricini2026} .

Yet, contemporary discourse relies on "gestalt analogies." While these broad comparisons successfully communicate a sense of historical magnitude, they fail to isolate the precise technical and social properties that carry the comparison. Is the printing press analogy a claim about the scale of dissemination, the disruption of existing knowledge authorities, or the speed of cultural change?  Each comparison implies a different historical lesson and a different governance posture but because gestalt analogies can collapse these distinctions, they can yield  governance mismatches: regulatory and design interventions that apply historical solutions to entirely different technical realities. 

This paper proposes a diagnostic framework to discipline how we construct and deploy historical analogies, thereby sharpening our capacity to analyze and govern the cultural impacts of emerging technologies. Drawing on ideas from the sociology of technology  \cite{Akrich1992,Winner1980,Hughes1987,Star1999,BowkerStar1999}, we decompose the interaction between technology and society into three coordinates:\begin{enumerate}
    \item \textbf{Epistemic Site}: The localized domain of human practice where the intervention occurs 
    \item \textbf{Governing Logic}: The organizing principle by which the technology constitutes and acts upon its object at that site  
    \item \textbf{Technical Mechanism}: The material or computational substrate through which the governing logic is concretely instantiated 
\end{enumerate}
The use of this framework is diagnostic: it serves as an apparatus for specifying what is being claimed when one technology is compared to another—for example, arguing that two technologies act on the same site under different logics (the index and the LLM at the site of discovery), or under the same logic via different mechanisms (the card catalog and the search engine).    

Crucially, this framework enables a distinguishing between two kinds of consequences.  Structural consequences are those produced by the mechanism itself; the technology cannot operate without making these consequences possible. Contingent consequences depend on choices that remain genuinely open For the printing press, the structural consequence was the mechanical ability to produce thousands of perfectly identical texts, permanently eliminating the drift and error of scribal hand-copying. Contingent consequences, by contrast, depend on institutional conditions and human choices that remain open. The printing press did not inherently fracture the Catholic Church; that impact was a contingent consequence, dependent on specific vernacular translation movements and the political protection of local princes\cite{eisenstein1980printing}.
 
We demonstrate the utility of this framework across two sites where generative AI is reshaping knowledge. First, at the site of information discovery, we compare the indexical logic of the catalog with the probabilistic inference of language models. Second, at the site of knowledge synthesis, we compare the editorial authority of the encyclopedia with the statistical consensus of GenAI. 

Ultimately, this paper does not aim to settle debates over the ultimate social utility of generative AI. Instead, it offers a rigorous method for organizing our claims about its impact. By moving from rhetorical gestures to diagnostic analogies, we show how policymakers and designers can locate the precise levers of governance that gestalt analogies leave obscured.

\section{A Framework for Diagnostic Analogy}
\textbf{ At what epistemic site does the intervention occur?} Technologies do not act on human practice as an undifferentiated whole; they intervene at particular sites of human activity. Identifying the site is crucial because the same technology can produce radically different outcomes depending on where it intervenes. Consider the telegraph \cite{Carey1983}: at the site of news transmission, it compresses the time between event and report, enabling the decontextualized factual dispatch. Yet at the site of market coordination, that same mechanism reorganized commodity trading and birthed modern financial speculation. An analogy that fails to specify the site risks missing the variegated impacts of the technology.

\textbf{By what governing logic does it organize its object at that site?} Two technologies can act on the same site under radically different organizing principles. The governing logic specifies the principle by which a technology organizes its object at the site of intervention—what it treats knowledge as, how it makes that object operable, what relationships between user, system, and content it presupposes. Naming the logic distinguishes technologies that share a site but produce different epistemic postures, and it makes visible the principle by which the technology constitutes its object.

\textbf{Through what technical mechanism is the logic being concretely instantiated?} The third coordinate specifies the concrete operation through which the governing logic is instantiated: the material and computational substrate, the procedural arrangement, the architecture by which the logic does its work. A single governing logic can be instantiated by many mechanisms. Historically, indexicality has been realized through many mechanisms in the back-of-book index, the library card catalog and PageRank\cite{Blair2010, Duncan2021, Krajewski2023}. All preserve the relationship between pointer and pointed-to while operating through entirely different material substrates. Separating logic from mechanism allows us to isolate which consequences follow from a technology's organizing principle versus its specific substrate, requiring what Bowker and Star \cite{BowkerStar1999} call an ``infrastructural inversion", a deliberate analytical turn toward the internal arrangements of the technology itself.

\section{Historical Analogies: From Rhetoric to Rigor}
\subsection{Information Discovery: From Indexicality to Inference} 

 Information discovery—locating specific knowledge within a larger epistemic landscape—has historically operated through the governing logic of indexicality \cite{Duncan2021, Krajewski2023}. The technology points to where knowledge lives, remaining structurally distinct from the thing indexed. A word directs you to a page; a URL directs you to a document. Because every output is a discrete source with an address, author, and institutional home, provenance is built into the architecture. You can always follow the arrow back.

Large language models (LLMs) operate under a different governing logic. They do not locate knowledge within a discrete object; they derive a probable answer from statistical patterns distributed across a training corpus. Knowledge is not retrieved but produced by the inferential act itself. The mechanism instantiating this logic—tokenization and next-token prediction—treats the corpus as a continuous statistical field. The model operates on sub-word fragments and their probability distributions, blending patterns across the corpus before any answer is produced. The distance between information and source, which indexicality preserves, collapses inside the model prior to output. While both the index and LLMs are discovery technologies, the index acts as a pointer to a bounded corpus the user will engage with, whereas LLMs acts as a generator, presuming a corpus the user will never see. 
 
 The framework identifies the erosion of provenance as structural because it follows from the mechanism of probabilistic inference itself: the model cannot generate its outputs without first dissolving the discreteness of its sources. Even with retrieval-augmented generation (RAG) systems, the generation of a fluent, unified answer from retrieved passages reintroduces the dissolution that the retrieval step was designed to prevent.  The source is consulted but the output is a probabilistic transformation of it. The user encounters the transformation. RAG shifts provenance erosion from total to partial, but it does not restore indexical architecture, and the primary epistemic encounter remains with inference, not source. 
Retrofitting citation tools onto generative outputs does not restore indexical architecture. Such tools impose an indexical logic—the pointer, the address, the source—onto a system whose architecture has already dissolved what the pointer would point to.  
The provenance problem thus is not an interface failure to be patched but a structural consequence of inference operating on a tokenized corpus. Addressing it requires intervention at the level of mechanism: alternative architectures that preserve indexical structures alongside generative ones, hybrid systems that maintain auditable knowledge pathways, or institutional requirements that prevent the complete displacement of indexical knowledge systems by inferential ones.

What cultural consequences can  cascade from this shift in mechanism and logic?  First, the slow erosion of provenance: the user no longer encounters knowledge as something with a traceable origin but as something delivered in finished form. Second, the disappearance of the lateral encounters of knowledge discovery where  one can stumble on the ‘adjacent’ knowledge object.  Inferential systems, optimized to return what the user asked for, foreclose the structural condition under which one stumbles into this \cite{Ginzburg1989} . Third is the flattening of context. In indexical architectures, a claim arrives marked by its origin: a Jane Austen sentence can be read differently from a Reddit post on the same theme  because the institutional and authorial context of the utterance is preserved and allows the user to understand these words differently Generative systems, by design, smooth across contexts. The same proposition, originating in a peer-reviewed paper or a chat forum, can surface in identical prose, indistinguishable to the user.
Ultimately, the user is repositioned from an active navigator weighing credibility to a passive recipient of finished answers. Under indexicality effective use required  choosing between competing results, weighing credibility, triangulating across conflicting claims, encountering context the user had not requested. Critics have argued that the move from card catalogue to search engine already eroded the deeper forms of this navigation, shortening attention and flattening reading \cite{Carr2010, Vaidhyanathan2012}. But in general even the indexical regime of the search engine kept the user in the position of moving toward knowledge. Under probabilistic inference, the technology has already performed the act of discovery on the user's behalf. The system produces finished answers where the previous system produced directions.

\subsection{Knowledge Synthesis: From Editorial Authority to Statistical Consensus}

Knowledge synthesis is the act of compressing a field of distributed claims into a consolidated artifact that a reader can consult in place of the field itself. The encyclopedia is one of the most systematic technologies ever built for this purpose, and its governing logic is editorial authority \cite{Yeo2001,Darnton1979}. Authority is derived from the scholar’s ability to judge what is significant and what is not, what needs to be synthesized and what should not. Synthesized claims in the encyclopedia are authoritative because they have been produced by identifiable human agent operating within accountable discursive communities whose judgments can be traced, challenged, and revised.  Diderot's editorial decisions in the first encyclopedia about what to include and how to frame it were visible, contested, and historically recoverable \cite{Darnton1979}. The reader encounters a curated compression of a field, and the curation is legible: the contributor is named, the apparatus is visible, the editorial frame is contestable. Knowledge is compressed for the reader, but the compression is accountable.

LLMs performs synthesis under a different governing logic of statistical authority. The underlying logic is consensus in the statistical sense of a central tendency across a distribution, not in the deliberative sense of agreement reached through argument. A claim surfaces in the synthesized output not because it has been verified by an expert or ratified through editorial process but because it is a high-probability pattern in the corpus. Authority is no longer tied to human judgment exercised within an accountable community; it is an emergent property of the volume of the data.

This logic is instantiated through the mechanism of probabilistic aggregation: statistical compression of a training corpus through gradient descent, weighted by whatever RLHF preferences were applied, filtered by whatever post-training guardrails were tuned.  The opacity of generative synthesis follows from the mechanism of statistical aggregation itself: gradient descent over a corpus of the scale required to produce fluent generation cannot be made auditable in the way a signed entry is auditable. The system does not have contributors whose judgments can be traced; it has weights whose origins are distributed across the entirety of the training process. Several consequences cascade from this shift. First, authority is decoupled from accountability. In the encyclopedia regime, the two travel together: a signed entry is also a target for response. Obscuring the process that produced a claim required active effort—censorship, suppression, the destruction of records. LLMs invert the default. Diffusing accountability is now the baseline condition, and recovering something like contestability requires sustained institutional effort against a system that produces none of it on its own. The direction of effort has reversed.

Second, LLMs flattens dis-consensus. The same query may produce different answers on different runs, but the user does not see the model arguing with itself. They do not see which views were marginal in the training data and which dominant, what was weighted up and what filtered out, where the corpus was dense and where it was sparse. The smoothing properties of statistical synthesis manufacture an appearance of consensus that does not correspond to any deliberative process.

\section{Implications for Governance} The framework directs governance attention along two distinct tracks. Structural consequences—provenance erasure under probabilistic inference, the opacity of statistical aggregation, the collapse of legible exclusion in generative synthesis—follow from the mechanisms these systems use to do what they do. They cannot be removed by interface design, transparency mandates, or post-hoc disclosure requirements that leave the mechanism intact. Addressing them requires intervention at the level of mechanism itself: architectural alternatives that preserve indexical structures alongside generative ones, hybrid systems that maintain auditable knowledge pathways for high-stakes domains, regulatory requirements that prevent the wholesale displacement of indexical and editorial knowledge systems by inferential ones in critical infrastructure. For example, recognizing structural consequences of tokenization, we could move towards strict RAG architectures with generation constrained strictly to retrieved context and inferential models that could be  restricted to acting solely as  interfaces that query and summarize fixed, auditable, indexical databases. These are difficult interventions because they touch the architecture rather than the surface, but they are the only interventions that reach the level at which the consequence is produced.  

The line between structural and contingent, however, is itself contestable, and part of the framework's purpose is to make this contestation possible. Is the opacity of GenAI structural, i.e., a necessary consequence of statistical aggregation at scale or contingent on choices not to invest in mechanistic interpretability, not to publish training data, not to build auditing infrastructure? 
While we organize this paper around sites, these are not stable bounded actions but themselves have histories, and part of what a new technology does is reshape the site it acts on. Information discovery, as a coherent domain of practice, was constituted in part by the indexical infrastructures that served it; what counts as discovery is already being transformed by inferential systems that fuse finding with answering. Knowledge synthesis as a publicly legible activity was constituted in part by the encyclopedia and its institutional surround; the chatbot reframes synthesis as conversation rather than consultation. Governance directed at preserving ``the site as it was" may misread what the technology is doing. The harder and more honest task is governing the reshaping itself: deciding which features of the prior site are worth preserving, which are obsolete, and which can only be preserved by deliberate institutional effort against the drift of the new architecture.

Large technical systems begin as malleable artifacts and harden, over time, into taken-for-granted infrastructures whose assumptions become invisible precisely because they are everywhere \cite{BowkerStar1999,Hughes1987}. The defaults of generative AI—opacity, statistical consensus, the fusion of finding and knowing, the disappearance of the legible exclusion that made earlier syntheses contestable—are not yet infrastructure. They are still choices, even if many are choices that no one has explicitly made. The window in which they can be recognized as choices, and treated as such, will not remain open indefinitely. The framework offered here is one instrument for keeping the window open: a way of asking, of any proposed intervention, whether it reaches the level at which the consequence is produced or merely the surface at which it appears. Whether the institutional scaffolding required to govern these systems emerges in time depends in part on whether the present moment is recognized as the moment in which the defaults can still be argued with rather than inherited.

\bibliography{example_paper}

@article{Star1999,
  author    = {Star, Susan Leigh},
  title     = {The Ethnography of Infrastructure},
  journal   = {American Behavioral Scientist},
  year      = {1999},
  volume    = {43},
  number    = {3},
  pages     = {377--391},
  doi       = {10.1177/00027649921955326}
}

@book{BowkerStar1999,
  author    = {Bowker, Geoffrey C. and Star, Susan Leigh},
  title     = {Sorting Things Out: Classification and Its Consequences},
  publisher = {MIT Press},
  address   = {Cambridge, MA},
  year      = {1999},
  doi       = {10.7551/mitpress/6352.001.0001}
}

@article{Winner1980,
  author    = {Winner, Langdon},
  title     = {Do Artifacts Have Politics?},
  journal   = {Daedalus},
  year      = {1980},
  volume    = {109},
  number    = {1},
  pages     = {121--136},
  url       = {http://www.jstor.org/stable/20024652}
}

@incollection{Akrich1992,
  author    = {Akrich, Madeleine},
  title     = {The Description of Technical Objects},
  booktitle = {Shaping Technology / Building Society: Studies in Sociotechnical Change},
  editor    = {Bijker, Wiebe E. and Law, John},
  publisher = {MIT Press},
  address   = {Cambridge, MA},
  year      = {1992},
  pages     = {205--224}
}

@incollection{Hughes1987,
  author    = {Hughes, Thomas P.},
  title     = {The Evolution of Large Technological Systems},
  booktitle = {The Social Construction of Technological Systems: New Directions in the Sociology and History of Technology},
  editor    = {Bijker, Wiebe E. and Hughes, Thomas P. and Pinch, Trevor},
  publisher = {MIT Press},
  address   = {Cambridge, MA},
  year      = {1987},
  pages     = {51--82}
}

@article{Petricini2026,
  author    = {Petricini, Tiffany},
  title     = {The power of language: framing AI as an assistant, collaborator, or transformative force in cultural discourse},
  journal   = {AI \& Society},
  year      = {2026},
  volume    = {41},
  number    = {2},
  pages     = {1005--1017},
  month     = {2},
  doi       = {10.1007/s00146-025-02586-2},
  note      = {Published online: September 10, 2025}
}

@online{Smith2025,
  author    = {Smith, Brad},
  title     = {The golden opportunity for American AI},
  publisher = {Microsoft On the Issues},
  year      = {2025},
  month     = {1},
  day       = {3},
  url       = {https://blogs.microsoft.com/on-the-issues/2025/01/03/the-golden-opportunity-for-american-ai/}
}

@online{Wheeler2024,
  author    = {Wheeler, Tom},
  title     = {Gutenberg's message to the {AI} era},
  publisher = {The Brookings Institution},
  year      = {2024},
  month     = {7},
  day       = {16},
  url       = {https://www.brookings.edu/articles/gutenbergs-message-to-the-ai-era/},
  note      = {Commentary based on ``From Gutenberg to Google and on to AI'' (Brookings Press, 2024)}
}

@article{Simosetal2022,
  author    = {Simos, Manolis and Konstantis, Konstantinos and Sakalis, Konstantinos and Tympas, Aristotle},
  title     = {``{AI} {CAN} {BE} {ANALOGOUS} {TO} {STEAM} {POWER}'' or From the ``{Post-Industrial} {Society}'' {To} the ``{Fourth} {Industrial} {Revolution}'': An {Intellectual} {History} of {Artificial} {Intelligence}},
  journal   = {ICON: Journal of the International Committee for the History of Technology},
  year      = {2022},
  volume    = {27},
  number    = {1},
  pages     = {97--116},
  publisher = {International Committee for the History of Technology (ICOHTEC)},
  url       = {https://www.icohtec.org/wp-content/uploads/2022/09/27-1-97.pdf}
}

@article{Farrelletal2025,
  author    = {Farrell, Henry and Gopnik, Alison and Shalizi, Cosma and Evans, James},
  title     = {Large {AI} models are cultural and social technologies},
  subtitle  = {Implications draw on the history of transformative information systems from the past},
  journal   = {Science},
  year      = {2025},
  volume    = {387},
  number    = {6739},
  pages     = {1153--1156},
  month     = {3},
  doi       = {10.1126/science.adr7848},
  url       = {https://www.science.org/doi/10.1126/science.adr7848}
}

@book{Carr2010,
  author    = {Carr, Nicholas},
  title     = {The Shallows: What the Internet Is Doing to Our Brains},
  publisher = {W. W. Norton \& Company},
  address   = {New York, NY},
  year      = {2010},
  isbn      = {9780393072228}
}

@book{Vaidhyanathan2012,
  author    = {Vaidhyanathan, Siva},
  title     = {The Googlization of Everything (And Why We Should Worry)},
  edition   = {1st},
  publisher = {University of California Press},
  address   = {Berkeley, CA},
  year      = {2012},
  pages     = {280},
  isbn      = {9780520272897}
}

@book{Yeo2001,
  author    = {Yeo, Richard},
  title     = {Encyclopaedic Visions: Scientific Dictionaries and Enlightenment Culture},
  publisher = {Cambridge University Press},
  address   = {Cambridge, UK},
  year      = {2001},
  doi       = {10.1017/CBO9780511497438},
  isbn      = {9780521651912}
}

@book{Darnton1979,
  author    = {Darnton, Robert},
  title     = {The Business of Enlightenment: A Publishing History of the {Encyclopédie}, 1775--1800},
  publisher = {Harvard University Press},
  address   = {Cambridge, MA},
  year      = {1979},
  isbn      = {9780674087859}
}

@article{Carey1983,
  author    = {Carey, James W.},
  title     = {Technology and Ideology: The Case of the Telegraph},
  journal   = {Prospects},
  year      = {1983},
  volume    = {8},
  pages     = {303--325},
  publisher = {Cambridge University Press},
  doi       = {10.1017/S036123330000371X},
  url       = {https://www.cambridge.org/core/journals/prospects/article/technology-and-ideology-the-case-of-the-telegraph/9C32F047433D0E498722E5248CD31B51},
  note      = {Published online by Cambridge University Press: 30 July 2009}
}

@book{Blair2010,
  author    = {Blair, Ann M.},
  title     = {Too Much to Know: Managing Scholarly Information before the Modern Age},
  publisher = {Yale University Press},
  address   = {New Haven, CT},
  year      = {2010},
  isbn      = {9780300165395},
  url       = {https://yalebooks.yale.edu/book/9780300165395/too-much-know}
}

@book{Duncan2021,
  author    = {Duncan, Dennis},
  title     = {Index, A History of the: A Bookish Adventure from Medieval Manuscripts to the Digital Age},
  publisher = {W. W. Norton \& Company},
  address   = {New York, NY},
  year      = {2021},
  isbn      = {9781324002543},
  note      = {UK edition published by Allen Lane (London) under the title ``Index, A History of the: A Bookish Adventure''}
}

@book{eisenstein1980printing,
  title={The printing press as an agent of change},
  author={Eisenstein, Elizabeth L},
  volume={1},
  year={1980},
  publisher={Cambridge University Press}
}

@book{Ginzburg1989,
  author    = {Ginzburg, Carlo},
  title     = {Clues, Myths, and the Historical Method},
  translator = {Tedeschi, John and Tedeschi, Anne C.},
  publisher = {Johns Hopkins University Press},
  address   = {Baltimore, MD},
  year      = {1989},
  isbn      = {9780801834585},
  note      = {Originally published in Italian as ``Miti, emblemi, spie: forme di identificazione e di discorso'' (Einaudi, 1986)}
}

@book{Krajewski2023,
  author     = {Krajewski, Markus},
  title      = {Paper Machines: About Cards \& Catalogs, 1548--1929},
  translator = {Krapp, Peter},
  publisher  = {The MIT Press},
  address    = {Cambridge, MA},
  year       = {2023},
  month      = {12},
  pages      = {224},
  series     = {History and Foundations of Information Science},
  isbn       = {9780262550857},
  note       = {Originally published in German as ``Zettelwirtschaft: Die Geburt der Karteikarte aus dem Geist der Verwaltung'' (Kulturverlag Kadmos, 2002); English hardcover edition published by MIT Press in 2011}
}

@misc{ng2017electricity,
  author = {Ng, Andrew},
  title = {Artificial {Intelligence} is the {New} {Electricity}},
  howpublished = {YouTube video, uploaded by Stanford Graduate School of Business},
  month = feb,
  year = {2017},
  note = {Talk given at the Stanford MSx Future Forum on January 25, 2017. \url{INSERT_YOUTUBE_URL_HERE}},
}
\bibliographystyle{icml2026}

\end{document}